\documentclass[conference]{IEEEtran}

\makeatletter

\usepackage{listings}

\def\ps@IEEEtitlepagestyle{%
  \def\@oddfoot{\mycopyrightnotice}%
  \def\@evenfoot{}%
}
\def\mycopyrightnotice{%
  {\footnotesize XXX-X-XXXX-XXXX-X/XX/\$XX.00~\copyright~20XX IEEE\hfill}
  \gdef\mycopyrightnotice{}
}

\usepackage{blindtext}
\usepackage{eso-pic}
\IEEEoverridecommandlockouts
\usepackage{cite}
\usepackage{amsmath,amssymb,amsfonts}
\usepackage{algorithmic}
\usepackage{graphicx}
\usepackage{textcomp}
\usepackage{xcolor}
\usepackage{booktabs} 
\usepackage{siunitx}
\usepackage{multirow} 
\usepackage{url} 
\usepackage{hyperref}  
\usepackage{nicefrac}
\usepackage{subfigure}
\graphicspath{{media/}} 
\usepackage{fancyhdr}  

\def\BibTeX{{\rm B\kern-.05em{\sc i\kern-.025em b}\kern-.08em
    T\kern-.1667em\lower.7ex\hbox{E}\kern-.125emX}}
    
\usepackage{eso-pic}
\newcommand\AtPageUpperMyright[1]{\AtPageUpperLeft{%
 \put(\LenToUnit{0.17\paperwidth},\LenToUnit{-2cm}){%
     \parbox{0.9\textwidth}{\raggedleft\fontsize{8}{11}\selectfont #1}}%
 }}%
\newcommand{\conf}[1]{%
\AddToShipoutPictureBG*{%
\AtPageUpperMyright{#1}
}
}

\begin{document}
\title{\vspace*{1cm} Detection and Measurement of Hailstones with Multimodal Large Language Models}

\author{\IEEEauthorblockN{Moritz Alker, David C. Schedl, and Andreas Stöckl}
\IEEEauthorblockA{\textit{Digital Media Lab} \\
\textit{University of Applied Sciences Upper Austria}\\
Hagenberg, Austria \\
\{s2410629001@students.fh-hagenberg.at, david.schedl@fh-hagenberg.at, andreas.stoeckl@fh-hagenberg.at\}}
}

\maketitle
\conf{\textit{  International Conference on Electrical and Computer Engineering Researches (ICECER 2025) \\ 
6-8 December 2025, Madagascar}}
\begin{abstract}
This study examines the use of social media and news images to detect and measure hailstones, utilizing pre-trained multimodal large language models. 
The dataset for this study comprises 474 crowd-sourced images of hailstones from documented hail events in Austria, which occurred between January 2022 and September 2024. These hailstones have maximum diameters ranging from 2 to 11\,cm. We estimate the hail diameters and compare four different models utilizing one-stage and two-stage prompting strategies. The latter utilizes additional size cues from reference objects, such as human hands, within the image.  
Our results show that pretrained models already have the potential to measure hailstone diameters from images with an average mean absolute error of 1.12\,cm for the best model. 
In comparison to a single-stage prompt, two-stage prompting improves the reliability of most models. 
Our study suggests that these off-the-shelf models, even without fine-tuning, can complement traditional hail sensors by extracting meaningful and spatially dense information from social media imagery, enabling faster and more detailed assessments of severe weather events. 
The automated real-time image harvesting from social media and other sources remains an open task, but it will make our approach directly applicable to future hail events.
\end{abstract}



\section{Introduction}

Hailstorms represent one of the most economically significant severe weather phenomena, with annual damages exceeding \$10 billion in North America alone \cite{loomis2018hail}. Climate change projections indicate an increase in hailstorm intensity and frequency in many regions, driven by enhanced atmospheric instability and stronger updrafts in supercell thunderstorms \cite{soderholm2020quantifying, nature_hailstone_2024}. Accurate hailstone size estimation is critical for agricultural risk assessment, insurance evaluation, and meteorological model validation.

Traditional hail detection methods rely on sparse ground-based sensors such as hail pads and disdrometers, which provide limited spatial coverage and temporal resolution. Weather radar systems, while offering broader spatial coverage, suffer from beam attenuation and ground clutter effects that reduce measurement accuracy, particularly for smaller hailstones \cite{lainer2023drone}. These limitations create significant gaps in our understanding of hail size distributions and their spatial variability.

Crowd-sourced data from social media platforms represents an emerging and largely untapped resource for severe weather documentation. Previous studies have demonstrated that social media reports can provide orders of magnitude more hail observations than traditional sensor networks \cite{blair2012creating, barras2019experiences}. However, extracting quantitative measurements from unstructured social media imagery remains challenging due to varying image quality, viewing angles, and the absence of standardized reference objects.

Recent advances in multimodal large language models (MLLMs) have shown remarkable capabilities in visual understanding and reasoning tasks \cite{achiam2023gpt, anthropic2024claude}. These models can process both textual and visual information, potentially enabling automated extraction of hailstone measurements from crowd-sourced imagery. Unlike traditional computer vision approaches that require extensive training data and domain-specific architectures, MLLMs can leverage pre-trained knowledge to perform complex visual reasoning tasks in a zero-shot manner.

Our study utilizes a dataset of 474 hailstone images derived from the European Severe Weather Database (ESWD) \cite{eswd_2009}, covering documented hail events in Austria from January 2022 to September 2024 (detailed in Section~\ref{sec:dataset}). The dataset encompasses hailstones with maximum diameters ranging from 2 to 11\,cm (mean: 4.17$\pm$1.46\,cm), with ground-truth measurements provided at 0.5\,cm accuracy. Images were manually annotated for reference object presence and viewing distance for quantitative analysis.

We evaluate four state-of-the-art MLLMs: GPT-4o and GPT-4o-mini from OpenAI \cite{achiam2023gpt}, Claude-Sonnet 4 from Anthropic \cite{anthropic2024claude}, and Gemini 2.5 Flash Lite from Google \cite{team2023gemini}. Two prompting strategies are compared: (P1) direct diameter estimation and (P2) a two-stage approach that first identifies reference objects before size estimation (detailed in Section~\ref{sec:results}). 
We summarize our findings and discuss future work in Section~\ref{sec:conclusion} and revisit related literature in the next section (Section~\ref{sec:relatedwork}).

\section{Related Work}
\label{sec:relatedwork}

This section reviews the relevant literature across four key areas: the development of multimodal large language models and their capabilities, computer vision approaches for image analysis and object size estimation, specialized techniques for hailstone detection and measurement from imagery, and the emerging use of social media data for weather phenomena documentation. These research domains collectively establish the foundation for applying modern AI techniques to crowd-sourced hail data analysis.

\subsection{Multimodal Large Language Models}

Recent advances in multimodal large language models (MLLMs) have revolutionized the integration of visual and textual modalities in AI systems. Caffagni et al. \cite{caffagni-etal-2024-revolution} provide a comprehensive survey of recent multimodal LLMs, detailing how these models integrate visual and textual modalities with dialogue interfaces. Their work covers architectures, alignment strategies, and training techniques, evaluating performance across diverse tasks including visual grounding, image generation and editing, and visual understanding. This survey provides crucial context for understanding the capabilities and limitations of multimodal LLMs in image-based analysis tasks.

The introduction of OpenAI's GPT-4 \cite{achiam2023gpt} marked a significant milestone in multimodal AI development. This large-scale model accepts both image and text inputs, producing text outputs, and demonstrates unprecedented performance across various benchmarks, achieving human-level scores in many domains. GPT-4 exhibited remarkable emergent abilities, including writing coherent narratives about images and solving visual mathematics problems without explicit optical character recognition. In our study, we harness these vision-language capabilities to analyze the visual content of pictures of hailstones, sourced from social media and online sources.

\subsection{Multimodal Models for Image Analysis and Object Size Estimation}

The development of CLIP by Radford et al. \cite{radford2021learningtransferablevisualmodels} introduced a foundational approach to learning transferable visual representations from natural language supervision. By training on 400 million image-caption pairs using a simple pre-training task of predicting caption-image matches, CLIP produces robust image embeddings aligned with language representations. The model achieves near state-of-the-art accuracy on ImageNet classification tasks without task-specific fine-tuning, relying solely on natural language class descriptions. This zero-shot classification capability demonstrates how multimodal models can identify objects purely from textual cues. 

The Visual Question Answering (VQA) paradigm, pioneered by Antol et al. \cite{Antol_2015_ICCV}, established a framework for multimodal reasoning that directly relates to hailstone size estimation tasks. VQA models process an image and a natural language question to generate textual answers, requiring detailed image understanding and reasoning beyond simple captioning. 
More recently, open‑source instruction‑tuned multimodal models like LLaVA (Large Language‑and‑Vision Assistant) \cite{liu2023llava} and its successor, LLaVA‑NeXT \cite{liu2024llavanext}, have pushed this field forward by combining CLIP’s vision encoder with a conversational LLM to enable image‑conditioned dialogue and reasoning for complex tasks such as reasoning, OCR, and world‑knowledge. In the context of hailstone imaging, such VQA systems offer a compelling framework for reasoning across modalities to estimate size. They can leverage known reference objects or scale cues present in images.


\subsection{Determining Hailstone Size from Images}

Specialized techniques for hailstone detection and measurement from imagery have emerged as critical tools for meteorological research. The HailPixel \cite{soderholm2020quantifying} technique, which employs drone photogrammetry to capture post-storm hail imagery and applies convolutional neural networks for robust hailstone detection combined with edge-detection algorithms for precise size measurement. This semi-automated approach can catalog tens of thousands of hailstones per survey, significantly exceeding the capabilities of traditional point sensors such as hail pads. The method's effectiveness was demonstrated in an Argentine case study, where HailPixel's large sample size notably improved the accuracy of hail size distribution characterization, particularly for the largest hailstones.

Advanced deep learning approaches for hailstone analysis have been further developed by Lainer et al. \cite{lainer2023drone}, who deployed Mask R-CNN object detection models on high-resolution drone imagery of hailstorm aftermath. Their system automatically identified and measured over 18,200 hailstones across a 750m² area during a 2021 Swiss severe storm event, producing detailed hail size distributions. This approach addresses fundamental limitations of point sensors, which sample only 0.2 m² areas, by mapping entire hail swaths and retrieving comprehensive hail size distributions for weather radar validation. The work exemplifies state-of-the-art AI applications in meteorological measurement and demonstrates the potential accuracy achievable through computer vision techniques for estimating hailstone dimensions.

Satellite-scale hail detection capabilities have been explored by Liu et al. \cite{liu2024study}, who developed deep neural network models to detect hailstorms from Meteosat geostationary satellite images. By analyzing multispectral infrared imagery and incorporating hail reports from the ESWD, their model achieved high accuracy in identifying hail-bearing clouds. While focused on satellite-scale detection rather than individual stone measurement, this research demonstrates the broader applicability of AI techniques for hail recognition across different scales and platforms, supporting the feasibility of automated hail detection from various imaging sources.

\subsection{Social Media for Weather Phenomena}

The potential of social media data for enhancing meteorological observations has been demonstrated through several pioneering studies. Blair and Leighton \cite{blair2012creating} conducted groundbreaking research showing that public social media postings can dramatically improve hail observations compared to traditional sensor networks. Their analysis of a severe hailstorm in Wichita (September 2010) gathered 464 hail size data points over 648km² using social media and post-storm ground surveys, with photographic evidence accompanying 93\% of reports. Remarkably, 94\% of the reports originated from social media platforms, and their analysis revealed record-breaking giant hailstones up to 197mm in diameter that would have been missed in official records. This work illustrates the vast untapped potential of social media imagery for meteorological research and directly motivates the application of advanced AI techniques to extract quantitative measurements from crowd-sourced hail photography.

Large-scale crowdsourcing initiatives have further validated the value of citizen-contributed weather data. Barras et al. \cite{barras2019experiences} reported on MeteoSwiss's crowdsourcing initiative, where a smartphone application enabled public submission of hail reports. This effort collected over 50,000 hail reports across Switzerland from 2015 to 2018, creating an exceptionally dense dataset that bridges gaps in official radar-based hail algorithms by providing ground truth at high spatial resolution. The study addresses quality control challenges and demonstrates the value of such data for validating and improving hail detection models, thereby reinforcing the importance of leveraging non-traditional data sources to capture hail events that would otherwise go unreported.

Real-time social media monitoring for hail events has been explored by Pramono et al. \cite{pramono2022crowdsourcing}, who focused on a specific hail event in Surabaya, Indonesia, using Twitter posts as real-time sensors. Their work demonstrated how hail-related tweets containing both text and images could be collected and analyzed to detect hail occurrence and assess disaster impacts. The study highlights methodologies for filtering relevant social media content and extracting valuable information for situational awareness during extreme weather events. This research directly supports our approach of utilizing multimodal LLMs on social media data, confirming that substantial hail evidence exists on social platforms and can be systematically harvested for quantitative analysis.

\section{The Dataset}
\label{sec:dataset}
As a basis for our study, we used a dataset created and provided by the ESWD \cite{eswd_2009}, which is operated by the European Severe Storms Laboratory (ESSL). The dataset comprises 521 hail events documented in Austria over nearly three years, from January 2022 to September 2024. 
Each sample consists of numerous features such as time event, country, state, location, latitude, longitude, maximum hailstone diameter, links to web sources, and various other descriptive information. 
The maximum diameter is the most relevant data source for our experiments and is provided with an accuracy of 0.5 cm. 

\begin{figure}[htbp]
\centering
\subfigure[Hailstone from the vicinity with a hand as reference object.]{\includegraphics[height=2.7cm]{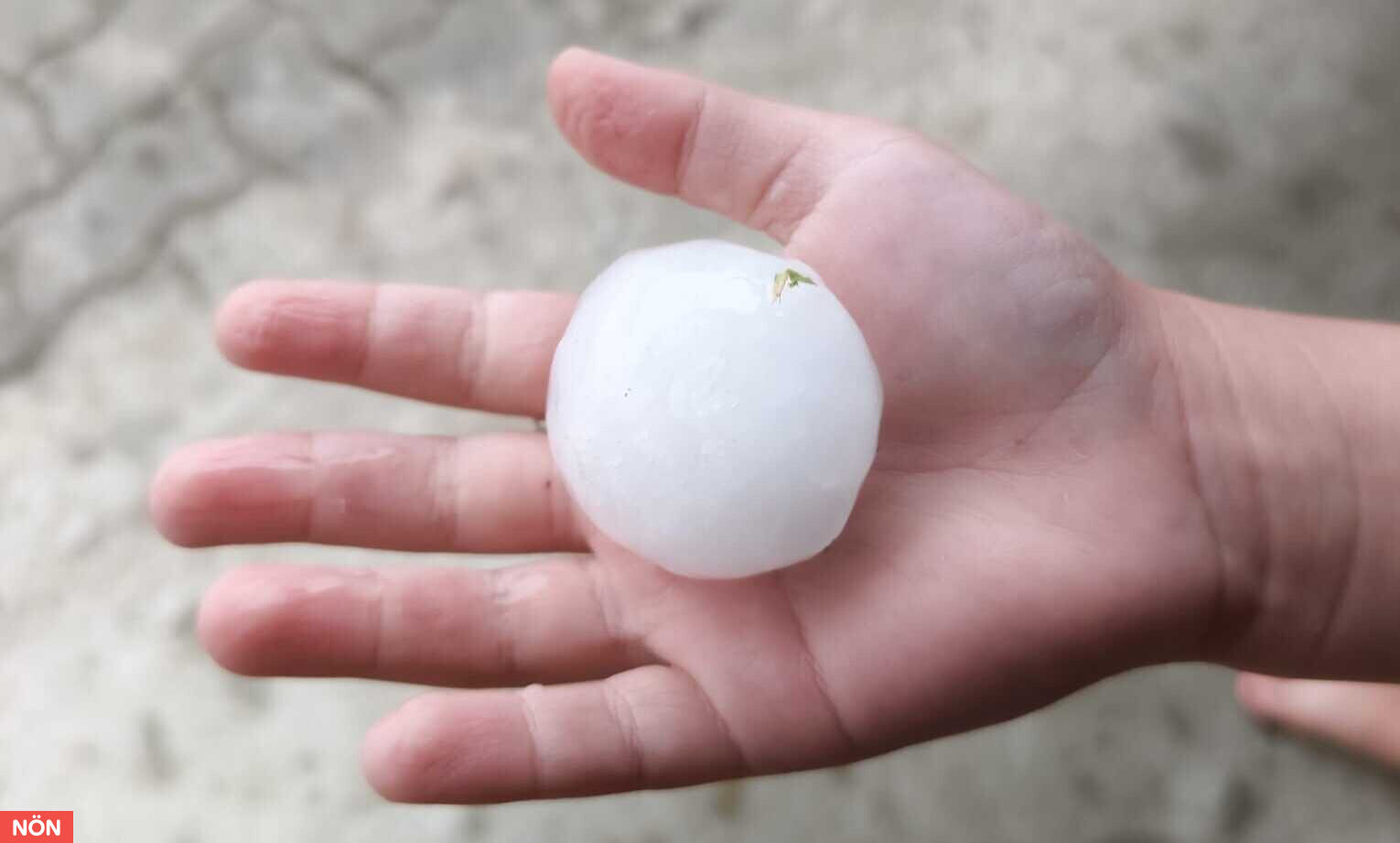}\label{fig:fig1}}
\quad
\subfigure[Hailstones from afar.]{\includegraphics[height=2.7cm]{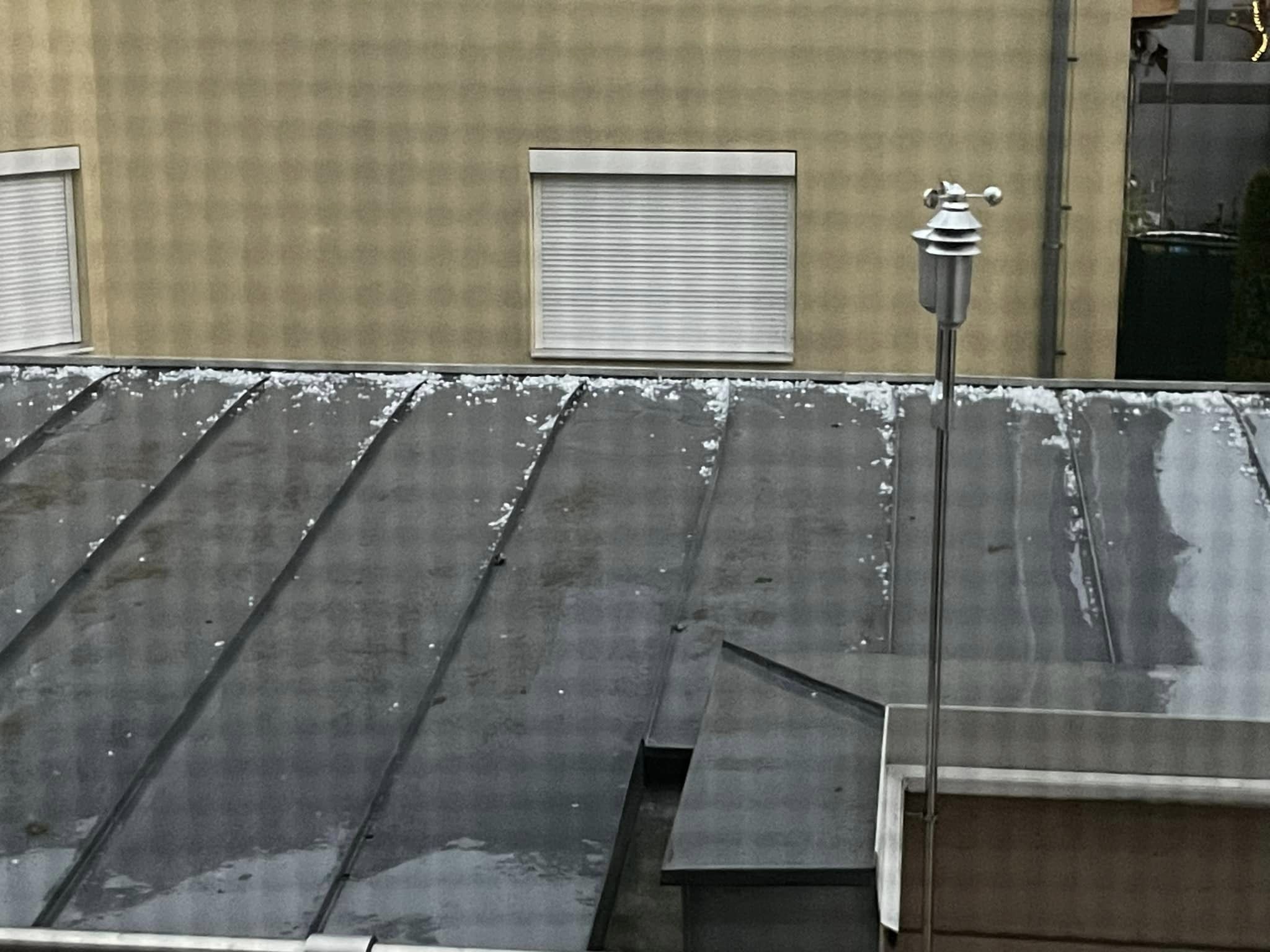}\label{fig:fig2}}
\caption{Examples of hailstone images from the dataset showing different viewing distances and reference objects.}
\label{fig:hailstone_examples}
\end{figure}

We prepared our dataset for model testing by extracting relevant events from the ESWD data, as each sample must contain at least one labeled maximum hailstone diameter and an image of the hailstones. We first removed all samples that did not include either a maximum hailstone diameter or a link to a source with an accessible image. Some events also contained multiple useful photos. We split the corresponding samples into multiple samples, resulting in a single image showing hail and the maximum diameter of a hailstone. Fig.~\ref{fig:hailstone_examples} shows example images from the dataset.

\begin{figure}[htbp]
  \centering
  \includegraphics[width=\linewidth]{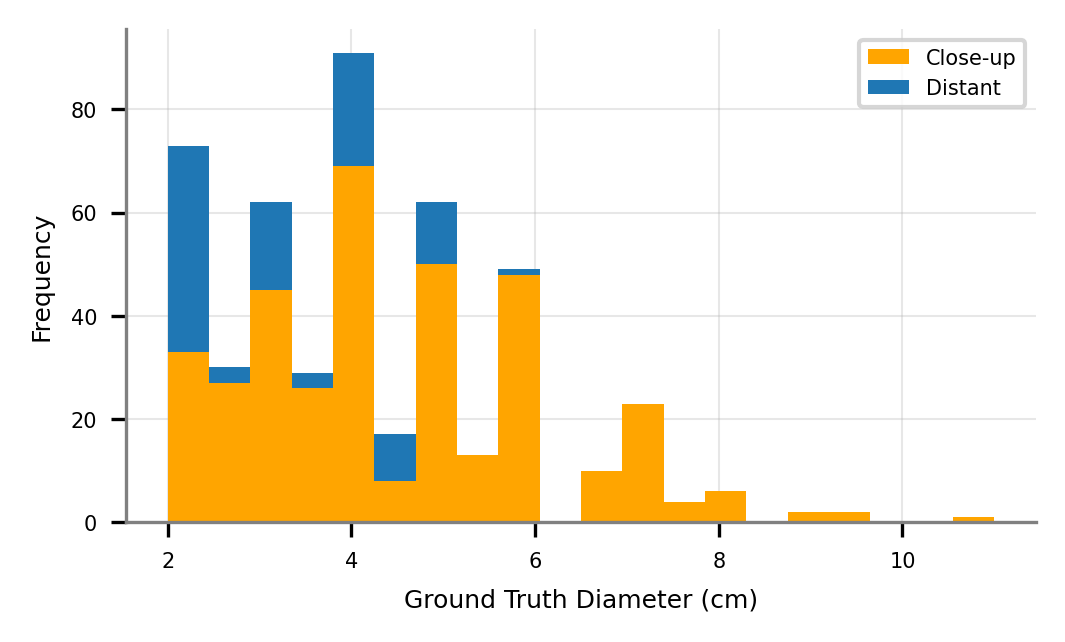}
  \caption{Histogram of 474 ground-truth hailstone diameters in our dataset. Differing colors indicate the distribution of close-up hailstones and hail in the distance.}
  \label{fig:hist}
\end{figure}

Furthermore, we manually added two new annotations that would be useful for evaluating the model's results. The first feature indicates whether there is a specific reference object in the image that might help to estimate the size of the hailstones---for example, a hailstone in a hand as shown in Fig.~\ref{fig:fig1}. 
In total, a single hand was the most common reference object, alongside hailstones, followed by no specific object, a ruler, and a coin, with 253, 136, 36, and 23 images, respectively. Furthermore, for some photos, multiple reference objects appeared, such as a hand holding a coin or ruler, while other images exhibited relatively uncommon objects, like tissue packs, cigarettes, or fruits. 
Our second additional annotation to each image is a binary feature indicating the distance from which the photo was taken, as the pictures are captured from varying distances: partly close-up, as in Fig.~\ref{fig:fig1}, and also from a distance, as in Fig.~\ref{fig:fig2}. In our dataset, 77.4~\% of images are close-up, while the remaining 22.6~\% show hail at a distance. 
After completing the data set preparation, we arrive at a total of 474 annotated images.
Ground-truth diameters ranged from $2$ cm to $11$ cm, with a mean of
$4.17\pm1.46$ cm and an inter-quartile range of 3 cm to 5 cm.
Figure~\ref{fig:hist} illustrates the strongly centered distribution; roughly
80\,\% of samples fall between 2 cm and 6 cm. 
Furthermore, most of the hailstones recorded from afar (distant) show relatively small hailstone diameters when compared to the close-up photos (cf. Fig.~\ref{fig:hist}). 

\section{Results}
\label{sec:results}

We tested four state-of-the-art multimodal-large-language models (MLLM). From OpenAI, we chose the GPT-4o (G4) and the GPT-4o-mini (G4m) models from the GPT-4 family \cite{achiam2023gpt}. 
From Anthropic \cite{anthropic2024claude}, we choose Claude-Sonnet 4 (CS4), and from Google we selected Gemini 2.5 Flash Lite (GFL) from the Gemini Models Family \cite{team2023gemini}. All models have vision capabilities and are suitable for our task. 
For unified access to all models, we relied on the LiteLLM Python library\footnote{{\url{https://github.com/BerriAI/litellm}}} with a maximum number of 100 output tokens. 
%
All models measured the maximum hail size of all 474 samples (see Section~\ref{sec:dataset}). We analyzed the accuracy of the results for all models by calculating the deviation of the predicted size from the actual size with two prompting strategies. 

\subsection{Prompting Strategies}

In prompt \textbf{P1}, the diameter is estimated directly with a single prompt (see excerpt from the prompt in \autoref{tab:prompts}). In the two-stage prompt \textbf{P2}, in a first step, a reference object (e.g., a hand, coin, ruler, lighter, etc.) is classified. The additional information related to the reference object with its typical dimensions is used as an aid in estimating size in the second step. For images without identifiable reference objects, contextual cues such as surrounding surfaces or environmental features are used to estimate the hailstone size (\autoref{tab:prompts}).
\begin{table}[htbp]
  \centering
  \caption{Prompting strategies used for hailstone diameter estimation.}
  \label{tab:prompts}
  \begin{tabular}{lp{6.3cm}}
    \toprule
    \textbf{Strategy} & \textbf{Prompt} \\
    \midrule
    \textbf{P1} & What is the maximum diameter of the hailstones in this image? Answer only with the diameter in cm as a float number. \\
    \midrule
    \textbf{P2 (Step 1)} & I am a climate researcher who deals with hail. Analyze this image of hailstones and check whether there is a reference object that can be used to identify the size of the hailstones. Answer only with one word. If you cannot recognize a reference object that is suitable, your answer has to be 'unspecified'. Examples: hand, coin, ruler, lighter. \\
    \midrule
    \multirow{3}{*}{\textbf{P2 (Step 2)}} & \textbf{For hand/coin/lighter:} Analyze the image and determine the maximum diameter of the hailstones using the \emph{[reference object]} as a reference. Use the known dimensions of \emph{[reference object and typical dimensions]} to estimate the hailstone diameter. Return only the estimated diameter as a float in centimeters without any text. \\
    \cmidrule{2-2}
    & \textbf{For ruler:} Analyze the image and determine the maximum diameter of the hailstones using the visible ruler as a reference. Directly measure the size of the hailstones from the markings on the ruler and return only the estimated diameter as float in centimeters without any text. \\
    \cmidrule{2-2}
    & \textbf{For unspecified:} Analyze the image and estimate the maximum diameter of the hailstones. Use contextual cues in the image, such as surrounding objects, surfaces, or environmental features to approximate the diameter of the hailstones. Return only the estimated diameter as float in centimeters without any text. \\
    \bottomrule
  \end{tabular}
\end{table}

\begin{figure}[htbp]
  \centering
  \includegraphics[width=\linewidth]{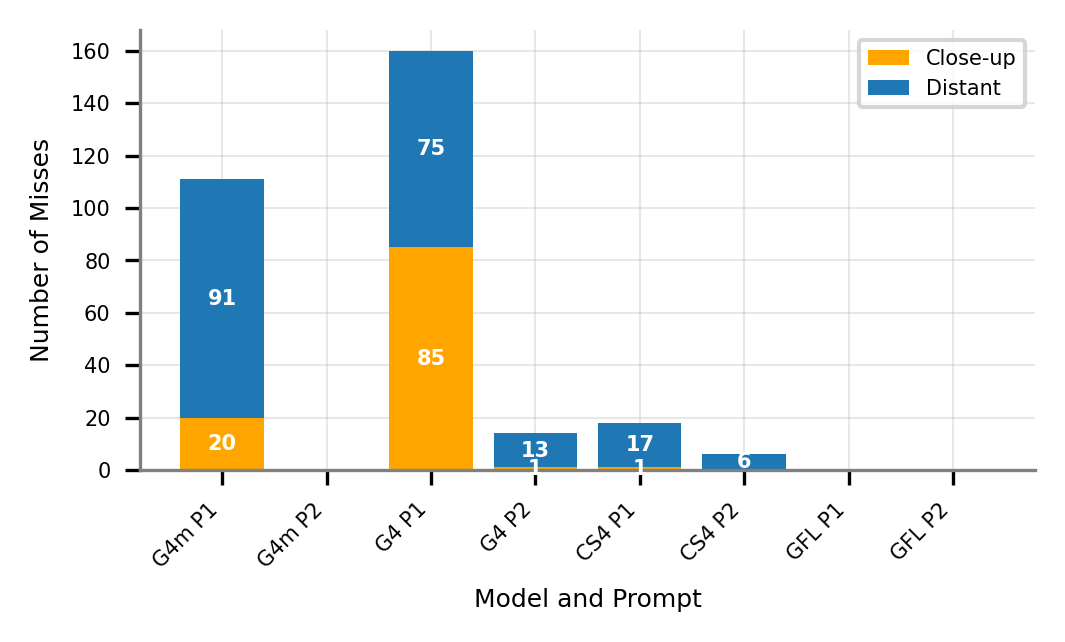}
  \caption{Histogram showing the number of misses per model and prompt.}
  \label{fig:misses}
\end{figure}

In both cases, the final expected answer of the prompting strategies was a floating-point number, which we rounded to 0.5~cm accuracy. 
Although the requested answer is a single number, the MLLMs sometimes generate extensive text. This is especially true for model CS4, as it began to \textit{think and reason} about measurements. For example, CS4 P1 responded with "Looking at the hailstones in the palm of the hand, I can use the hand as a reference for scale. [\ldots]". In most cases, however, we were still able to extract a meaningful measurement as a number was provided within the response. In such cases, we used the first numerical value for our analysis. 
If the model failed to provide a number, we set the estimated diameter to zero. 
Figure~\ref{fig:misses} and \autoref{tab:metrics} provide an overview of misses for the evaluated models and prompts. While the OpenAI models (G4 and G4m) show a high number of missed responses with prompt P1 (160 and 111, respectively), GFL always provided a numerical estimate for both prompts. Overall, there is a clear tendency for P2 to reduce the number of misses compared to P1. Furthermore, the majority of response misses are caused by distant hailstones in the images. 
Also, note that the Anthropic API (used for model CS4) failed to process six images due to encoding issues. 

\subsection{Model Performance}

Table~\ref{tab:metrics} summarises the error statistics for the vision-language models with corresponding prompting strategies. 
Model \textbf{G4} with strategy \textbf{P2} achieved the lowest mean-absolute error (MAE $=1.12$ cm) and the highest Pearson correlation with the ground truth ($r=0.71$).  
All models exhibited a negative bias, indicating a systematic underestimation of the actual hailstone size.  
The P2 prompting variants reduced MAE on average by 18.6\,\% compared with their P1 prompts.

\begin{table}[htbp]
  \centering
  \caption{Error metrics for each model. Boldface highlights the best score
  per column.}
  \label{tab:metrics} 
  \begin{tabular}{lrrrrr}
    \toprule
    Model & MAE (cm) & RMSE (cm) & Bias (cm) & $r$ & Miss \\
    \midrule
    G4 P2        & \textbf{1.12} & \textbf{1.47} &    -0.72 & \textbf{0.71} &     14 \\
    CS4 P1       &     1.17 &     1.53 &    -0.76 &     0.65 &     18 \\
    G4m P2       &     1.20 &     1.56 & \textbf{-0.49} &     0.52 & \textbf{0} \\
    CS4 P2       &     1.20 &     1.59 &    -0.93 &     0.63 &      6 \\
    GFL P1       &     1.28 &     1.70 &    -0.89 &     0.60 & \textbf{0} \\
    GFL P2       &     1.47 &     1.89 &    -1.20 &     0.51 & \textbf{0} \\
    G4m P1       &     1.65 &     2.21 &    -0.87 &     0.49 &    111 \\
    G4 P1        &     2.04 &     2.82 &    -1.63 &     0.39 &    160 \\
    \bottomrule
  \end{tabular}
\end{table}

\begin{figure}[htbp]
  \centering
  \includegraphics[width=\linewidth]{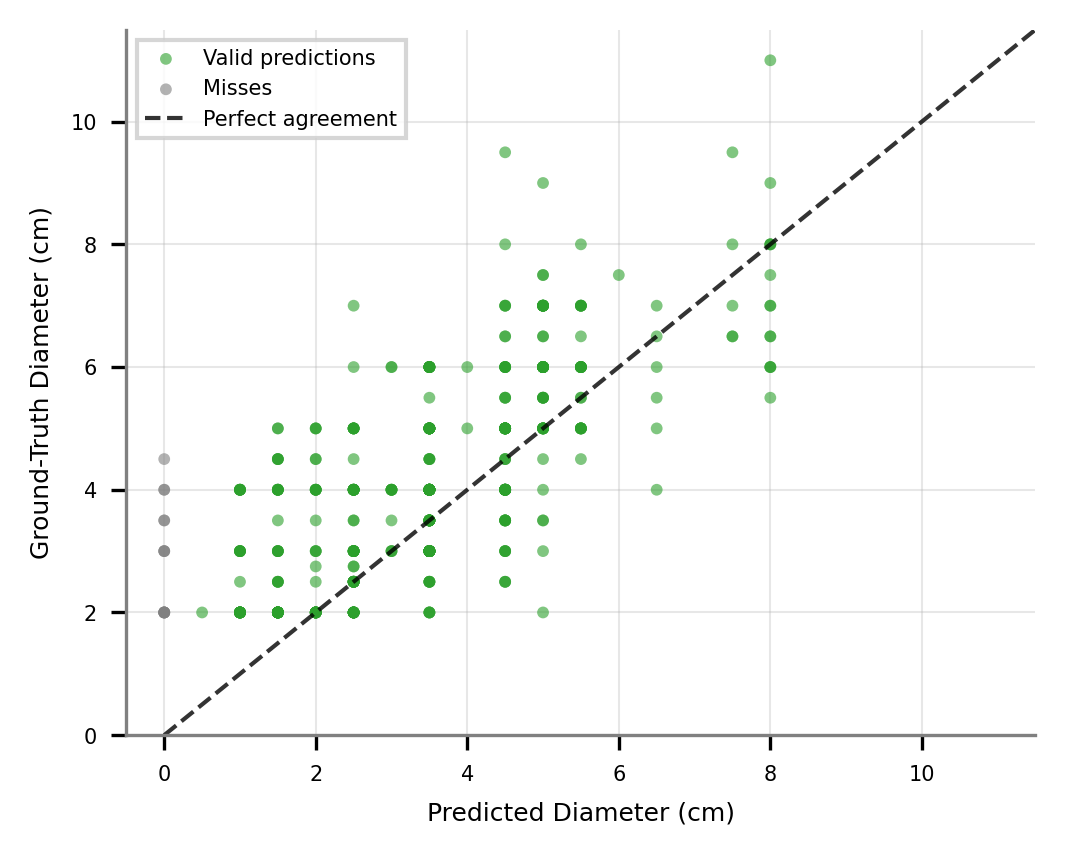}
  \caption{Ground-truth versus G4 P2 estimates.  The dashed line
  denotes perfect agreement.}
  \label{fig:scatter}
\end{figure}

The performance of the best model and prompt (G4 P2) is visualised in
Fig.~\ref{fig:scatter}.  Points scatter predominantly above the identity line, confirming the overall negative bias of  $0.72$\,cm on average.  

\subsection{Effect of the Reference Object}

\begin{figure}[htbp]
  \centering
  \includegraphics[width=0.9\linewidth]{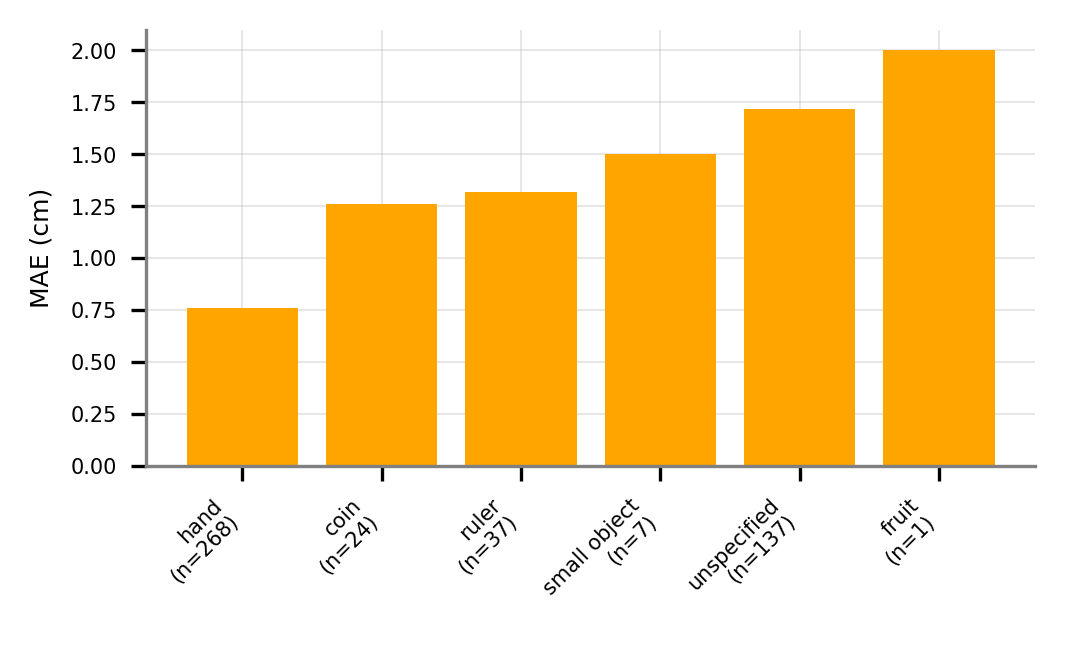}
  \caption{Mean-absolute error (MAE) of G4 P2 grouped by reference object.}
  \label{fig:refbar}
\end{figure}

Reference objects markedly influenced estimation accuracy
(Table~\ref{tab:refobj}; Fig.~\ref{fig:refbar}).  Images containing a \emph{human hand} provided the strongest scale cue, yielding an MAE of $0.75$ cm.  In contrast, photographs without an explicit scale (\emph{unspecified/other})
nearly doubled the error ($1.73$ cm).  Although line-rulers offer an absolute scale, their advantage was diminished (MAE $=1.32$ cm), presumably because rulers were often tilted, introducing perspective distortions.

\begin{table}[htbp]
  \centering
  \caption{G4 P2 MAE by reference object.}
  \label{tab:refobj}
  \begin{tabular}{lrr}
    \toprule
    Reference object & $n$ & MAE (cm) \\
    \midrule
    Hand                    & 268 & \textbf{0.75} \\
    Coin / Bottle cap       &  24 & 1.26 \\
    Ruler                   &  37 & 1.32 \\
    Small household object  &   7 & 1.50 \\
    Unspecified / Other     & 137 & 1.73 \\
    Fruit\textsuperscript{*}&   1 & 2.00 \\
    \bottomrule
  \end{tabular}\\
  \textsuperscript{*}Single sample; result not generalizable.
\end{table}

\section{Conclusions}
\label{sec:conclusion}

This study demonstrates that \emph{off-the-shelf} multimodal large language models (MLLMs) can already extract quantitative information about hailstones from user-generated imagery with surprising accuracy. On a crowd-sourced test set of 474 annotated photographs, the best-performing model, \textbf{GPT-4o} (abbreviated as G4), achieved a mean absolute error of \SI{1.12}{\centi\meter} and a correlation of $r = 0.71$ with ground truth measurements using a two-stage prompting strategy (P2), as shown in Table~\ref{tab:metrics}. While all evaluated models exhibited a systematic underestimation of hailstone size (with an average bias of $-0.7$\,cm), the two-stage prompting reduced the overall error by 18.6\,\% compared to single-stage prompts (P1). Furthermore, the number of missed responses was reduced from a total of 289 to 20 by using a two-stage approach. 

In summary, the following insights emerge from the evaluation. The presence of clear reference objects—particularly human hands—substantially improved accuracy, reducing the error to 0.75\,cm, while the absence of scale cues nearly doubled it (Table~\ref{tab:refobj}). This highlights the importance of interpretable, planar scale information in crowd-sourced images. The underestimation bias was consistent across models, indicating shared limitations in interpreting three-dimensional scales from two-dimensional inputs. This may stem from a conservative tendency of the models when faced with visual ambiguity. 
Surprisingly, the GPT-4o-mini model (G4m) showed the smallest bias of $-0.49$\,cm.

\subsection*{Impact}

These findings show that current MLLMs, even without fine-tuning, can complement traditional hail sensors by extracting meaningful and spatially dense information from social media imagery. Such models have the potential to become valuable tools in operational meteorology, enabling faster and more detailed assessment of severe hail events. With modest improvements, they could help address the growing socio-economic risks posed by climate-driven increases in hailstorm severity.

\subsection*{Limitations and Future Directions}

The dataset used in this study is limited to hail events from 2022 to 2024 in Austria, which may limit the generalizability of our findings to regions with different environmental conditions or social media conventions. Manual annotations of distance classes and reference objects may introduce subjective bias. Furthermore, the study does not yet incorporate automated, real-time image harvesting from social media—an essential step for practical nowcasting applications.

Future work should prioritize expanding the dataset to increase geographic and contextual diversity. To address the underestimation bias, exploring geometric priors or perspective correction techniques could prove valuable. For operational use, the most critical next step is developing an automated, real-time pipeline that harvests images from social media, filters for relevance and quality. It integrates the size estimates into meteorological nowcasting systems. Such a system would fully realize the potential of MLLMs for high-resolution monitoring of severe weather.

\section*{Acknowledgments}

We would like to thank the European Severe Storms Laboratory (ESSL) and Thomas Schreiner for providing the dataset.

\bibliographystyle{unsrt}  
\bibliography{references}

\end{document}